\documentclass[runningheads]{llncs}
\usepackage[T1]{fontenc}
\usepackage{amssymb, amsmath, amsfonts}
\usepackage{hyperref}

\usepackage{multirow}
\usepackage{graphicx,verbatim}
\usepackage{color}

\begin{document}
\title{Tractography-Driven Synthetic Data Generation for Fiber Bundle Segmentation in Tracer Histology}
\titlerunning{Tractography-Driven Synthetic Data for Tracer Histology}
%
\author{Kyriaki-Margarita Bintsi \inst{1} \and
Sparsh Makharia \inst{2} \and
Ya\"{e}l Balbastre \inst{3} \and
Joselyn Romero Avila \inst{4} \and
Julia F.~Lehman \inst{5} \and
Suzanne N.~Haber \inst{5,6} \and
Anastasia Yendiki \inst{1} 
}
%

\authorrunning{K. M. Bintsi et al.}
%
\institute{Athinoula A. Martinos Center for Biomedical Imaging, Massachusetts General Hospital and Harvard Medical School, Charlestown, MA, United States 
\email{kbintsi@mgh.harvard.edu}\\
\and
Krea University, Sri City, Andhra Pradesh, India
\and
Department of Experimental Psychology, University College London, London, United Kingdom
\and
Universidad Nacional Mayor de San Marcos, Lima, Peru
\and
Department of Pharmacology and Physiology, University of Rochester School of Medicine, Rochester, NY, United States
\and
McLean Hospital, Belmont, MA, United States}


  
\maketitle              
\begin{abstract}
Diffusion MRI (dMRI) tractography enables non-invasive reconstruction of white-matter pathways, but its accuracy is fundamentally limited by indirect, low-resolution measurements of axonal organization. Tracer injection studies in non-human primates provide a gold standard for validating dMRI tractography. This, however, requires time-consuming manual annotation of fiber bundles in histology sections. We propose a synthetic-data augmented framework for automated fiber bundle segmentation in macaque tracer histology. Our approach uses ex vivo dMRI tractography as a generative prior to synthesize 2D image patches for training. This provides us with sufficiently realistic foreground texture, which we compose with backgrounds from blockface photos and diversify via domain randomization. A 2D U-Net is trained on mixed real and synthetic patches. Experiments on held-out brains demonstrate improved generalization across brains and fiber bundle densities compared to training with real data only. Training with synthetic data only leads to poor performance, underscoring the need for real supervision. Overall, our approach achieves performance comparable to the state-of-the-art while requiring 3x less manually annotated data.


\end{abstract}
%
%

\section{Introduction}
Diffusion MRI (dMRI) tractography enables non-invasive reconstruction of white matter pathways in health and disease~\cite{catani2006diffusion,yang2021diffusion}. However, it uses indirect measurements of axonal orientations based on water diffusion and has millimeter-scale resolution, orders of magnitude coarser than individual axons. Anatomic tracer studies in non-human primates (NHPs) offer a gold standard for validating tractography~\cite{safadi2018functional,yendiki2022post}, as they allow direct visualization of the complete trajectory of axonal projections at single axon resolution~\cite{schilling2019limits,jbabdi2013human}. Yet reconstructing fiber pathways from tracer histology requires labor-intensive manual annotation, severely limiting the amount of tracer data available for validation studies.

Machine learning can automate this process, but progress has been constrained by labeled data scarcity and variability across tracers, injection sites, and brains. Methods for axon segmentation in electron microscopy \cite{wei2021axonem,naito2017identification} or fluorescence microscopy in rodents and marmosets~\cite{winnubst2019reconstruction,friedmann2020mapping,yan2022mapping} do not generalize to macaque tracer data, which are particularly valuable for translational studies due to known human--macaque homologies~\cite{chatterjee2009estimating}. Initial efforts to deploy deep learning for segmenting fiber bundles in macaque tracer data~\cite{sundaresan2025self} demonstrated feasibility but relied on \textit{ad hoc} post-processing steps to reduce false positives. Recent work employed semi-supervised pre-training to improve accuracy and generalizability to sparse bundles, while obviating the need for post-processing~\cite{bintsi2025fully}, but still required manual annotations from multiple brains.

Synthetic training data, i.e., images generated from ground-truth segmentation labels, offer a promising strategy for reducing the burden of manually labeling real data. Domain randomization, i.e., systematically varying geometry, appearance, and noise during synthesis, has proven effective in encouraging models to learn structural patterns robust to dataset-specific artifacts, and has found applications across biomedical imaging domains~\cite{pezoulas2024synthetic,gopinath2024synthetic,chollet2024neurovascular}. In this approach, a network trained on synthetic data can learn to segment real data, not because the synthetic data are realistic, but because they are very diverse. To date, this strategy has not been explored for segmenting tracer histology.

Here, we present a synthetic-data augmented framework for fiber bundle segmentation in macaque tracer histology. Our synthesis framework aims to replicate the key feature of real tracer data, where a tracer labels axons projecting from a single injection site in the gray matter, as they leave the injection site, split into bundles of varying densities, and follow different paths through the white matter to reach their destinations. We propose to leverage dMRI tractography as a geometric prior to synthesize image patches with this appearance. Importantly, we do not need the end-to-end trajectory of the tractography streamlines to be anatomically correct, as we only use them to generate image patches. We compose the streamlines with backgrounds from blockface photos, diversified via domain randomization. Combined with limited real annotated sections, this achieves strong cross-brain generalization with substantially reduced annotation effort. This paper has the following contributions: (1) A tractography-driven synthesis pipeline with controllable domain randomization. (2) A mixed real+synthetic training strategy matching state-of-the-art~\cite{bintsi2025fully} performance with 3$\times$ less annotated data. (3) Ablations on the amount of real data, the synthetic-to-real data ratio, and randomization strength, showing that even a single annotated section yields competitive performance. The code is publicly available at \url{https://github.com/lincbrain/synthetic-tracer-histology-segmentation}.

\section{Methods}

\begin{figure}[t]
\includegraphics[width=\textwidth]{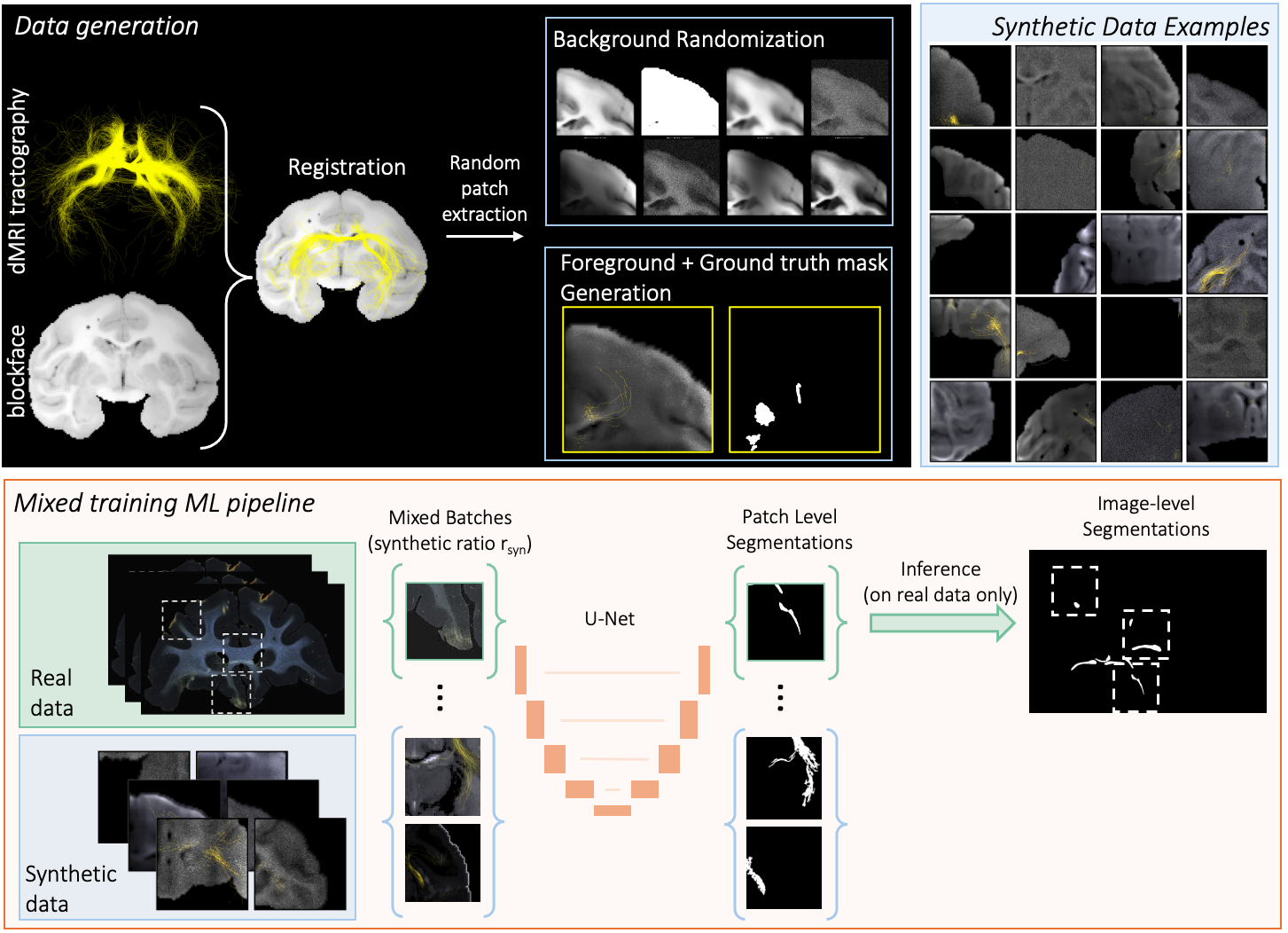}
\caption{Overview of the proposed pipeline. \textbf{Top:} Ex vivo dMRI is registered to a blockface image from the same brain, and tractography streamlines from a cortical ``injection site'' are mapped into blockface space. We use subsets of these streamlines with varying fiber densities as the foreground, and compose them with randomized backgrounds to produce synthetic image--mask pairs. \textbf{Bottom:} Real histology patches and synthetic patches are mixed in each mini-batch with synthetic ratio $r_{\text{syn}}$ to train a 2D U-Net. At inference, only real images are processed, and overlapping patch predictions are merged into full-section bundle segmentations.}
\label{fig:pipeline}
\end{figure}

\subsection{Synthetic Data Generation}
\label{sec:synthesis}
We design an on-the-fly synthesis pipeline that generates realistic image--mask pairs from a single ex vivo macaque brain (M0) for which both whole-brain dMRI and co-registered blockface photographs are available. The dMRI data were acquired on a 4.7T preclinical scanner using a 3D echo-planar sequence (0.5\,mm isotropic, 515 gradient directions, $b_{\max} = 40{,}000$\,s/mm$^2$), comprising three b-shells ($b = 4{,}000$, $8{,}000$, $12{,}000$\,s/mm$^2$) with 64, 64, and 128 gradient directions respectively, following the protocol of~\cite{jones2020insight}. 
We used multi-shell multi-tissue constrained spherical deconvolution~\cite{dhollander2018feasibility} implemented in MRtrix3 and seeded probabilistic tractography (iFOD2 algorithm, maximum bending angle $30^\circ$, minimum FOD amplitude $0.1$, all other parameters at default) 10,000 times per voxel, repeated for 96 voxels distributed throughout the cortex to mimic different ``injection sites'', yielding anatomically diverse fiber configurations that effectively decouple the synthetic training distribution from any single brain's anatomy.

After sectioning, blockface photographs were captured at 16--21\,\textmu m in-plane resolution and the stacked volume was registered deformably to the $b{=}0$ dMRI volume using ANTs~\cite{avants2009advanced}. Let $B\in\mathbb{R}^{W\times D\times H}$ and $\mathcal{S}=\{s_1,\ldots,s_N\}$ denote the blockface volume and tractography streamlines transformed into blockface space, respectively. Each synthetic sample $(\hat{I},\hat{M})$ is constructed as follows.

\noindent\textbf{Background.}
To bridge the domain gap between blockface photography and dark-field microscopy, we sample a slice index $y^\ast \sim \mathcal{U}\{1,\ldots,D\}$ and extract the coronal section $B_{y^\ast}(x,z)=B(x,y^\ast,z)$.
We apply mild CLAHE, a power-law dark-field transform with
$\gamma \sim \mathcal{U}(0.8,1.2)$ restricted to the brain mask, and a
dark-field colormap~\cite{reza2004realization}, to emulate the appearance of a brain section.
To further bridge the domain gap, the image is processed with augmentation presets \cite{chollet2024neurovascular} sampled with weighted probability, spanning optical transforms, parametric noise models, smooth multiplicative bias fields, and Lab color perturbations. 
Finally, we sample a random $P\times P$ crop ($P=512$).

\noindent\textbf{Foreground (synthetic tracer signal).}
Given the slice plane $y^\ast$, we select streamlines originating from the current
``injection site'' that pass within distance $\tau$ of the plane:
\begin{equation}
\mathcal{S}_{y^\ast} = \{s_i \in \mathcal{S} : \min_{(p_x, p_y, p_z) \in s_i} |p_y - y^\ast| \le \tau\},
\label{eq:streamline_selection}
\end{equation}
where $(p_x, p_y, p_z) \in s_i$ denotes a 3D point along streamline $s_i$, $p_y$ its $y$-coordinate (coronal axis), and $\tau$ is an adaptive thickness parameter scaled with output resolution.
Points on $\mathcal{S}_{y^\ast}$ are densified via linear interpolation at 0.2-voxel spacing and projected onto the coronal plane. We then sample a fraction $\rho_{\max}\sim \mathcal{U}(0.02,0.05)$ of streamlines, from the densely seeded tractography streamlines, and subsample further by $r_{\text{viz}}\sim \mathcal{U}(0.5,0.7)$ for rendering. 
Each streamline is rendered as an anti-aliased curve with stochastic linewidth $w\sim \mathcal{U}(0.5,1.5)$\,px, opacity fading linearly from $\alpha=0.4$ at $y^\ast$ to $0$ at distance $\delta$, and per-streamline color jitter in the orange--yellow range to mimic tracer appearance. 
Optionally ($p_{\mathrm{inj}}=0.3$), injection-site artifacts are added by rendering short curved streamlets radiating from a random point with saturated orange hues.
The background crop and rendered tracer signal are composited to form the synthetic image $\hat{I}$.

\noindent \textbf{Ground-truth mask.}
These binary masks mimic the appearance of fiber bundle annotations. They are generated from a dense streamline set, so that $\hat{M}$ captures the full spatial extent of each bundle even when the synthesized tracer signal is sparse.
We accumulate a density map $D(x,z)$ by rasterizing line segments between consecutive projected streamline points (Bresenham). We then Gaussian-smooth ($\sigma{=}2.0$\,px) and threshold using an
absolute count threshold $\theta{=}0.08$: 
\begin{equation}
\hat{M}_{\text{raw}}(x,z)=\mathbf{1}\Big[(G_{\sigma} * D)(x,z)
\ge \theta\Big].
\end{equation}
Post-processing applies per-component dilation within a disk (to expand bundles while preserving separation) and small-object removal. This decoupling, sparse fibers in $\hat{I}$ but dense bundles in $\hat{M}$, encourages the model to detect bundles even when only partially visible, reflecting real tracer histology.

\noindent \textbf{On-the-fly generation and domain randomization.}
Samples are synthesized dynamically during training: each mini-batch randomly selects a seed voxel, slice index, and patch location, rendering the corresponding sample through the full pipeline and providing large diversity without storage overhead (${\sim}4$\,s/patch on CPU). Synthesis parameters were selected empirically via visual inspection and prior knowledge of tracer appearance. Beyond the in-pipeline transforms, we apply a second augmentation stage during training that includes spatial transforms (flips, rotations $\pm 20^\circ$, elastic deformations), appearance transforms (color jitter, random gamma, hue--saturation shifts), and stochastic degradations (Gaussian/ISO noise, mild blur).

\subsection{Mixed Training Pipeline}
An overview of the pipeline is shown in Figure~\ref{fig:pipeline}. Each training batch combines real and synthetic samples at ratio $r_{\text{syn}}$. Real patches use foreground-aware sampling (half centered on fiber pixels, half at random locations) to address class imbalance. At inference, overlapping $1024{\times}1024$ patches are aggregated by averaging to produce full-section probability maps, followed by connected component filtering to remove small spurious detections smaller than 800\,pixels..

\section{Experiments}
\noindent\textbf{Datasets.}
\label{sec:datasets}
We use high-resolution coronal histological sections from four in-house macaque datasets (M1--M4) with bidirectional tracer injections at multiple cortical sites. Sections (50\,\textmu m thick) were digitized at 0.4\,\textmu m in-plane resolution, with every 24\textsuperscript{th} section processed to visualize a specific tracer (1.2\,mm effective inter-slice spacing). This sampling strategy follows previous tracer studies in non-human primates \cite{lehman2011rules,haynes2013organization}. Expert neuroanatomists annotated fiber bundles (i.e. groups of axons traveling in close proximity from injection sites to terminal fields) into three density classes: dense, moderate, and sparse. Approximately 30 sections per brain were annotated, yielding ${\sim}90$ section-level evaluation samples across the three held-out brains (M2--M4), each downsampled $16{\times}$ to a ${\sim}4\text{k}{\times}4\text{k}$  pixel field of view at 6.4\,\textmu m effective resolution.
We train using annotations from a single brain (M1). Generalization is evaluated on three held-out brains: M2 and M3 (partially used to train the SOTA, precluding direct comparison on these brains) and M4 (fully held-out, enabling comparison with all methods).

\noindent\textbf{Evaluation Metrics.}
\label{sec:metrics}
Standard overlap metrics (Dice, IoU) are ill-suited to this task: severe class imbalance deflates their values, and the exact boundaries of each bundle are less important than its location and extent. We therefore adopt the component-based evaluation of~\cite{sundaresan2025self}: connected component matching yields \textit{True Positive Rate (TPR)} per density class (dense/moderate/sparse), \textit{False Discovery Rate} $\text{FDR} = \text{FP}/(\text{TP}+\text{FP})$, and \textit{average TP/FP per section}.

\noindent\textbf{Baselines.}
\label{sec:baselines}
SAM2~\cite{ravi2024sam} and MedSAM2~\cite{zhu2024medical} are evaluated as zero-shot baselines, prompted automatically from the white-matter region. Both perform poorly on tracer histology, occasionally detecting dense bundles but mostly missing moderate and sparse bundles, and FDR close to 1.0, underscoring the need for task-specific training. We further compare against: (i)~\textbf{Sundaresan et al.}~\cite{sundaresan2025self}, the initial self-supervised deep-learning method for this task, which relies on cross-section post-processing; (ii)~\textbf{Bintsi et al.}~\cite{bintsi2025fully}, the current state-of-the-art (SOTA), which uses semi-supervised pre-training and requires annotations from three brains (M1--M3); and (iii)~a \textbf{real-only} baseline using the pipeline from \cite{bintsi2025fully}, trained on only one brain (M1) instead of three, to isolate the effect of synthetic data from the effect of additional annotated brains.

\noindent\textbf{Implementation Details.}
\label{sec:implementation}
We use a 2D U-Net~\cite{ronneberger2015u} with nine encoder--decoder levels, each comprising two $3{\times}3$ convolutions with instance normalization and LeakyReLU. Training uses AdamW~\cite{loshchilov2017fixing} with learning rate $10^{-4}$, cosine annealing with linear warmup, batch size~8, mixed-precision (FP16), and a combined BCE--Dice loss for 150 epochs. Each batch mixes real and synthetic data at ratio $r_{\text{syn}} = 0.7$ (default). Real patches ($1024{\times}1024$) are sampled with 50\% centered on fiber pixels to address class imbalance; synthetic patches ($512{\times}512$) are generated on-the-fly and resized to match. Real data receives conservative augmentation (flips, small rotations), while synthetic data undergoes extensive domain randomization (Sec.~\ref{sec:synthesis}). Training runs on a single NVIDIA RTX 6000 GPU.

\section{Results}

\begin{figure}[t]
\includegraphics[width=\textwidth]{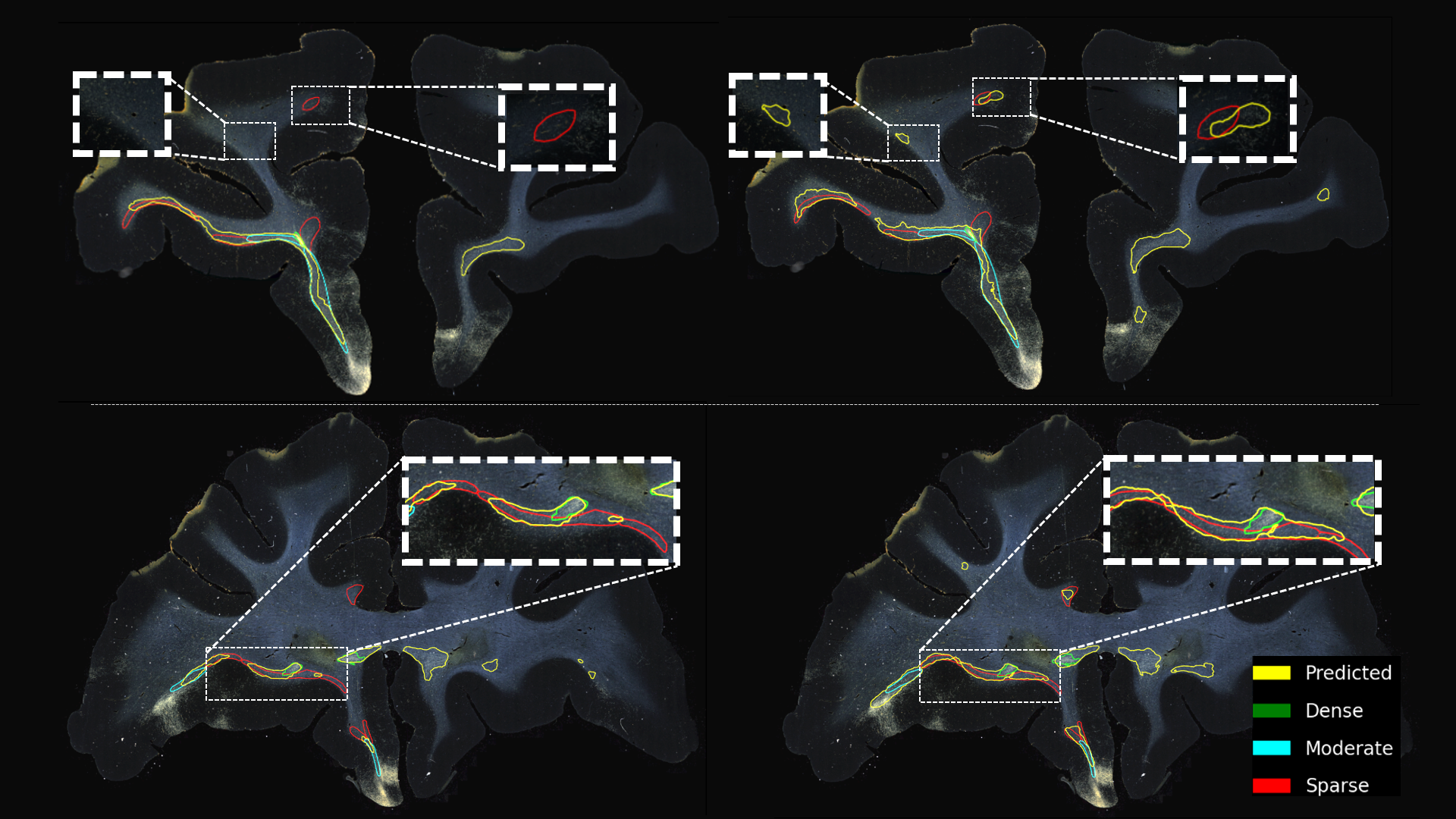}
\caption{Visual comparison between the \textbf{state-of-the-art method~\cite{bintsi2025fully} (left)} and \textbf{our approach (right)} on two histological sections from the held-out brain M4. Ground-truth bundles are outlined by density class (dense: green, moderate: cyan, sparse: red); predictions are shown in yellow. Insets highlight regions where the baseline often misses sparse bundles that our model detects, consistent with the higher sparse-bundle TPR in Table~\ref{tab:main}, at the cost of more false positives. Annotations are provided only within the ipsilateral hemisphere; predictions outside this region are excluded from evaluation.}
\label{fig2}
\end{figure}

\noindent\textbf{Main comparison.}
Table~\ref{tab:main} compares all methods on the held-out test brain M4. Our method outperforms the SOTA~\cite{bintsi2025fully} in overall detection ($\text{TP}_\text{avg}$~3.24 vs.~3.00) and achieves similar sparse-bundle sensitivity (0.79 vs.~0.78), while being trained only on one brain instead of three. It substantially outperforms Sundaresan et al.~\cite{sundaresan2025self} ($\text{TP}_\text{avg}$~3.24 vs.~1.76) and improves sparse bundle detection over the real-only baseline (0.79 vs.~0.52), at the cost of a higher FDR (0.36 vs.~0.18).

Figure~\ref{fig2} shows qualitative results on two sections from M4, comparing the SOTA ~\cite{bintsi2025fully} (left) with our method (right). Both methods recover dense and moderate bundles well, but the SOTA baseline often misses sparse bundles that our model detects, in line with the higher sparse-bundle TPR in Table~\ref{tab:main}. These additional detections come at the cost of slightly more FPs.

\begin{table} [t]
\caption{
Performance comparison on the held-out test brain M4. We report true positive rates (TPR) for dense (D), moderate (M), and sparse (S) bundles, average true/false positives per section (TP$_\text{avg}$, FP$_\text{avg}$), and false discovery rate (FDR).
}
\centering
\small
\begin{tabular}{l|c|ccc|cc|c}
\hline
\textbf{Method} & \textbf{Train data} & \multicolumn{3}{c|}{\textbf{TPR} $\uparrow$} & \textbf{TP$_\text{avg}$} & \textbf{FP$_\text{avg}$} & \textbf{FDR} \\
 & (\# brains) & \textbf{D} & \textbf{M} & \textbf{S} & $\uparrow$ & $\downarrow$ & $\downarrow$ \\
\hline
Sundaresan et al.~\cite{sundaresan2025self} & 1 & 0.85 & 0.75 & 0.41 & 1.76 & 2.91 & 0.62 \\
Real-only~\cite{bintsi2025fully}                                    & 1 & 0.88 & 0.75 & 0.52 & 2.88 & 0.64 & 0.18 \\
Bintsi et al.~\cite{bintsi2025fully}         & 3 & 0.97 & 0.90 & 0.78 & 3.00 & 1.36 & 0.30 \\
Bintsi et al.\ (+ pretrain)~\cite{bintsi2025fully} & 3+10 & 0.97 & 0.83 & 0.62 & 3.10 & 0.64 & 0.20 \\
Proposed                                     & 1+synth & 0.97 & 0.89 & 0.79 & 3.24 & 2.06 & 0.36 \\
\hline
\end{tabular}
\label{tab:main}
\end{table}

\noindent\textbf{Cross-brain generalization.}
Since we train on M1 only, we can additionally evaluate on M2 and M3, which the SOTA~\cite{bintsi2025fully} used for training. Table~\ref{tab:cross_brain} shows that our mixed real+synthetic training (Proposed) generalizes well across held-out brains, improving over~\cite{sundaresan2025self} especially on moderate and sparse bundles (e.g., on M2, $\text{TPR}_\text{M/S}$ 0.95/0.64 vs.\ 0.85/0.48). Compared to the real-only baseline, the proposed model substantially increases $\text{TPR}_\text{S}$ (from 0.35 to 0.64 on M2 and from 0.38 to 0.62 on M3), at the cost of higher FP$_\text{avg}$ and FDR on M2 (0.40 vs.\ 0.11), reflecting a systematic shift toward higher recall under cross-brain shift. Notably, on M3 the FDR is reduced relative to the real-only baseline (0.37 vs.\ 0.47), suggesting that the elevated FDR on M2 reflects a challenging domain shift rather than a systematic failure mode.

\begin{table}[t]
\caption{
Cross-brain generalization on M2 and M3 (methods trainable on M1 only; 
the SOTA~\cite{bintsi2025fully} cannot be evaluated here as it trains on M1--M3).
}
\centering
\small
\begin{tabular}{l|l|ccc|cc|c}
\hline
\textbf{Brain} & \textbf{Method} & \multicolumn{3}{c|}{\textbf{TPR} $\uparrow$} & \textbf{TP$_\text{avg}$} & \textbf{FP$_\text{avg}$} & \textbf{FDR} \\
 &  & \textbf{D} & \textbf{M} & \textbf{S} & $\uparrow$ & $\downarrow$ & $\downarrow$ \\
\hline
\multirow{3}{*}{M2}
 & Sundaresan et al.~\cite{sundaresan2025self} & 0.92 & 0.85 & 0.48 & 2.06 & 3.23 & 0.61 \\
 & Real-only~\cite{bintsi2025fully}              & 0.65 & 0.77 & 0.35 & 1.60 & 0.23 & 0.11 \\
 & Proposed                                    & 0.98 & 0.95 & 0.64 & 2.73 & 3.10 & 0.40 \\
\hline
\multirow{3}{*}{M3}
 & Sundaresan et al.~\cite{sundaresan2025self} & 0.96 & 0.76 & 0.27 & 0.89 & 3.63 & 0.79 \\
 & Real-only~\cite{bintsi2025fully}                & 0.86 & 0.81 & 0.38 & 1.86 & 2.00 & 0.47 \\
 & Proposed                                    & 0.95 & 0.95 & 0.62 & 3.93 & 3.06 & 0.37 \\
\hline
\end{tabular}
\label{tab:cross_brain}
\end{table}

\begin{table}[t]
\caption{Ablation studies (averaging over M2--M4). Proposed denoted with $^{*}$.}
\centering
\footnotesize
\setlength{\tabcolsep}{2.2pt}
\renewcommand{\arraystretch}{1.05}
\begin{minipage}[t]{0.475\linewidth}
\centering
\begin{tabular}{l|c|c}
\hline
\textbf{Training data }& \textbf{TPR (D/M/S)} & \textbf{FDR} \\
(\# of sections) &  $\uparrow$ & $\downarrow$\\
\hline
1       & 0.88/0.80/0.57 & 0.32 \\
5     & 0.80/0.71/0.53 & 0.33 \\
30 (1 brain)$^{*}$  & 0.97/0.93/0.69 & 0.38 \\
\hline
\end{tabular}
\end{minipage}
\begin{minipage}[t]{0.475\linewidth}
\centering
\begin{tabular}{l|c|c}
\hline
\textbf{$r_{\text{syn}}$} & \textbf{TPR (D/M/S)}$\uparrow$ & \textbf{FDR}$\downarrow$ \\
\hline
0\%              & 0.93/0.91/0.69 & 0.44 \\
30\%             & 0.96/0.89/0.60 & 0.37 \\
50\%             & 0.94/0.90/0.64 & 0.42 \\
70\%$^{*}$    & 0.97/0.93/0.69 & 0.38 \\
100\%            & 0.64/0.36/0.26 & 0.94 \\
\hline
\end{tabular}
\end{minipage}
\label{tab:ablations}
\end{table}

\noindent\textbf{Ablations studies.}
\noindent\textit{Real data amount.} Table~\ref{tab:ablations} (left) reports mean performance across held-out brains M2--M4. A single annotated section paired with synthetic data already achieves competitive sensitivity ($\text{TPR}_\text{D/M/S}$ 0.88/0.80/0.57) with the lowest FDR (0.32); using a full annotated brain further improves detection, especially for sparse bundles (0.69 vs.\ 0.57), at the cost of a higher FDR (0.38). Performance is non-monotonic so 5~sections do not consistently outperform 1, indicating high variance in the low-shot regime.

\noindent\textit{Synthetic ratio.} Table~\ref{tab:ablations} (right) varies the proportion of synthetic images per batch, averaging over M2--M4. Synthetic-only training fails (mean $\text{TPR}_\text{S}$ 0.26, FDR 0.94), confirming that real supervision is essential. Compared to the real-only setting ($r_{\text{syn}}{=}0\%$), mixed real+synthetic training improves dense and moderate TPR, maintains sparse recall, and reduces FDR, achieving a better overall precision--recall trade-off. This holds across a broad range of synthetic ratios ($r_{\text{syn}}\in\{30\%,50\%,70\%\}$), indicating robustness to this hyperparameter.

\noindent\textit{Domain randomization.} 
Across M2--M4 we also compare our default domain randomization to a less aggressive variant. Default DR consistently achieves a better precision--recall trade-off: on all brains it maintains comparable TPR while reducing FDR (e.g., 0.40 vs.\ 0.58 on M2 and 0.36 vs.\ 0.49 on M4), suggesting that stronger augmentation acts as an effective regularizer across brains.

\section{Discussion and Conclusion}
We have presented a mixed real+synthetic training strategy for fiber bundle segmentation in tracer histology that leverages dMRI tractography as a generative prior. Our results demonstrate that synthetic augmentation effectively complements limited real supervision: performance comparable to the SOTA~\cite{bintsi2025fully} is achieved with annotations from a single brain (vs.\ three), and even a single annotated section yields competitive detection when paired with synthetic data. Crucially, synthetic data alone is insufficient. Real supervision remains essential to anchor the learned representations but at intermediate mixing ratios, synthesis acts as a regularizer that improves precision relative to real-only training.

The primary limitation of our approach is elevated FPs on some held-out brains, which we attribute to domain shifts in background appearance, tracer characteristics, and injection patterns not fully captured by the current synthesis pipeline; notably, on M3 FPs increase while FDR \textit{decreases} relative to the real-only baseline, indicating that TP gains outweigh FP increases. We note that this does not reflect circularity from the tractography prior: our model learns only \textit{local} axon-bundle texture from thin histological sections, not long-range connectivity, so the dominant failure mode of tractography does not propagate into the segmentation task. Mixed training with real annotated histology further grounds the model in biologically valid tracer appearance. Increasing training diversity through additional real brains or improved synthesis would further improve generalization. The non-monotonic behavior in the low-shot regime also suggests that which sections are annotated matters as much as how many, motivating active selection strategies. Future work will focus on improved synthesis to better capture tracer-dependent intensity 
profiles and staining artifacts via generative models~\cite{friedrich2024deep}, and uncertainty-aware filtering to reduce FPs.

Ultimately, by reducing dependence on annotated data through synthesis, this work enables reliable bundle segmentation not only for similar protocols but also for unseen tracers and tissue preparations, thus making a key step toward high-throughput mesoscale connectivity mapping.


    

\begin{credits}
\subsubsection{\ackname} 
This work was supported by the center for Large-scale Imaging of Neural Circuits (LINC), an NIH BRAIN Initiative Connectivity across Scales (CONNECTS) comprehensive center (UM1-NS132358). Additional support was provided by the National Institute for Mental Health (R01-MH045573, P50-MH106435) and the National Institute for Neurological Disorders and Stroke (R01-NS119911, R01-NS127353).

\subsubsection{\discintname}
The authors have no competing interests to declare that are
relevant to the content of this article. 
\end{credits}

%
%
\bibliographystyle{splncs04}
\bibliography{Paper-4001}
\end{document}